\documentclass[final]{fctr}
\usepackage{algorithm}
\usepackage{algpseudocode}
\usepackage{doi}

\newcommand{\num}[1]{#1}
\makeatletter
\@ifundefined{compactenum}{%
  \newenvironment{compactenum}{\begin{enumerate}[itemsep=0pt,topsep=2pt,parsep=0pt,leftmargin=1.6em]}{\end{enumerate}}%
}{}
\makeatother
\title{ICJ-Graph: A Citation Graph and a Layered Document Corpus of the International Court of Justice}

\author{%
  Jirameth Wannasiwaporn$^{1\dagger}$\quad Pawitsapak Akarajaradwong$^{2}$\quad Chompakorn Chaksangchaichot$^{2}$\\[3pt]
  \affil{$^{1}$Computer Engineering and Digital Technology, Chulalongkorn University, Thailand\\$^{2}$FourCorners Team, VISAI AI, Thailand}\\[1pt]
  \email{6633033021@student.chula.ac.th, \{pawitsapaka\_visai, chompakornc\_pro\}@vistec.ac.th}\\[3pt]}

\begin{document}
\maketitle
\begingroup
\renewcommand{\thefootnote}{\fnsymbol{footnote}}
\footnotetext[2]{This work was conducted during Jirameth
Wannasiwaporn's internship at VISAI AI.}
\endgroup

\begin{abstract}
We present ICJ-Graph, a machine-readable resource connecting documents of the International Court of Justice (ICJ) to cited cases, legal provisions and United Nations resolutions.
Court decisions, written pleadings and oral records form distinct document layers, preserving the distinction between the Court's reasoning and the parties' submissions.
The collection contains \num{6{,}396} ICJ documents and \num{505} documents from other courts and legal bodies, together with \num{12{,}375} provisions and \num{20{,}291} UN resolutions.
The data files contain text, metadata and embeddings.
The accompanying pipeline lets users load the graph and generate embeddings for use in their own retrieval systems.
To illustrate the resource's use, we evaluate five models on citation prediction for \num{199} decisions using documents available before each decision.
Tools designed for citation research improve citation-prediction F1 over answering without tools for all five models across case, provision and resolution citations.
\end{abstract}
\section{Introduction}
\label{sec:intro}
The International Court of Justice (ICJ) is the principal judicial organ of the United Nations and its jurisprudence is a central object of study in international law and empirical legal scholarship~\citep{posner2005biased,alschner2018selfcitation}. Existing machine-readable resources, however, cover only part of the Court's public record.
The Corpus of Decisions of the ICJ (CD-ICJ)~\citep{fobbe2023cdicj} contains judgments, orders, advisory opinions and appended individual opinions, but excludes the written pleadings and oral records that document the parties' arguments.
These materials record the parties' submissions alongside the Court's decisions.

The relevant materials extend beyond the Court's own documents to the cases, legal provisions and UN resolutions they cite.
Existing UN resources provide resolution texts and citation networks for the General Assembly and Security Council~\citep{Mesquita_2024,fobbe_2025_15191558}.
Our aim is to connect these types of material within a single resource for research on the ICJ.

We present \textbf{ICJ-Graph}, a layered corpus connecting ICJ decisions, written pleadings and oral records to cited cases, legal provisions and UN resolutions.
The pipeline converts the source PDFs to structured text, segments each document according to its own organization, embeds the resulting passages and resolves citations to cases, provisions and UN General Assembly or Security Council resolutions.

We evaluate the resource through citation prediction over \num{199} target decisions.
For each target, the model receives only case information available before the decision date, while the graph hides the target, same-day material, later documents and undated documents.
We compare five models answering without tools, using general graph queries and text search, or using tools designed for citation research.
The task separately scores citations to cases, provisions and UN resolutions over three repeated runs.

Our contributions are:
\begin{compactenum}
    \item A machine-readable corpus with \num{6{,}396} ICJ documents organized into decision, pleadings and oral layers, alongside \num{505} documents from other courts and adjudicative bodies.
    \item A citation graph connecting documents to cases, legal provisions and UN resolutions, with \num{1{,}206{,}400} document passages and \num{169{,}617} resolution passages embedded for semantic retrieval.
    \item A citation-prediction benchmark covering \num{199} decisions and five models, with retrieved documents restricted to dates before each decision.
\end{compactenum}

\section{Related Work}
\label{sec:related}

\paragraph{ICJ and international-court resources.}
CD-ICJ \citep{fobbe2023cdicj} is the fullest of these, a machine-readable collection of the
Court's published decisions; CD-PCIJ \citep{fobbe2022cdpcij} is its counterpart for the
predecessor Permanent Court. Both stop at decisions. \citet{alschner2018selfcitation}
construct a self-citation network and \citet{posner2005biased} study judicial voting, both
annotating metadata our graph can carry rather than releasing the record. The iCourts
database \citep{icourtsdb} is a searchable multi-court portal with downloadable citation
networks, but a query interface rather than a bulk, openly licensed corpus and states no
reuse terms.

\paragraph{UN resolution resources.}
\citet{Mesquita_2024} release a corpus of General Assembly resolutions adopted between 1946 and 2019 with a network of citations between resolutions.
The Corpus of Resolutions: UN Security Council (CR-UNSC)~\citep{fobbe_2025_15191558} provides resolution texts, metadata, draft texts and meeting records, together with citations to Security Council and General Assembly resolutions.
These resources establish UN resolutions and their citation links as material for computational research.
ICJ-Graph connects resolution records to citations in court documents alongside links to cases and legal provisions.

\paragraph{Legal citation networks.}
A line of work ranks precedents by their position in the citation graph.
\citet{fowler2008authority} introduced it for the US Supreme Court and
\citet{bellutta2024icjnetwork} apply it directly to the ICJ, reporting that reverse
PageRank outperforms hub and authority scores for identifying influential decisions.
\citet{lupu2012precedent} caution that citation centrality can reflect strategic rather
than legal weight. Beyond international courts, \citet{akarajaradwong2026fourcorners} build a Thai legal knowledge graph and find structured graph retrieval beats vector search.

\paragraph{Legal judgment prediction.}
Predicting a case's outcome from its text is an established task.
\citet{aletras2016ecthr} popularized it for the European Court of Human Rights (ECtHR) and \citet{medvedeva2020using} scaled it up, showing that accuracy drops sharply under temporal evaluation, where a model is tested on cases decided after the ones it trained on; \citet{medvedeva2023rethinking} distinguish post-hoc \emph{identification} from genuine \emph{forecasting}. 
Surveys~\citep{cui2022survey} catalogue many datasets but none targeting international-court precedent selection. 
\citet{chalkidis2023chatgpt} find that instruction-tuned LLMs underperform smaller fine-tuned models on legal prediction despite strong legal-knowledge recall.

ICJ-Graph brings decisions, written pleadings and oral records together with the cases, provisions and UN resolutions cited in those documents.
Its contribution is a connected resource that preserves the document layers and identifies the source documents for citation links.

\section{The Corpus and Citation Graph}
\label{sec:corpus}

We represent the corpus as one graph.
Cases, documents, provisions, UN resolutions, judges and countries are nodes and edges encode the citations and relations among them.
Document citations link to cases, instruments, provisions and UN resolutions. Provisions and resolutions can also cite others of their own type.

\subsection{Background of ICJ Data}
An ICJ case is a proceeding on the Court's General List with a record of the Court's decisions and the parties' submissions.
Decisions include judgments, orders and advisory opinions together with the separate and dissenting opinions appended by individual judges.
The parties' submissions appear in written pleadings such as memorials and in oral records that transcribe the hearings.
The graph schema in \S\ref{subsec:icj_schema} preserves these document types and connects them to the cases, legal provisions and UN resolutions they cite.

Article 38(1)(a) and (d) of the ICJ Statute guides our inclusion of treaties and judicial decisions~\citep{icj_statute}.
These provisions recognize treaties as a source of applicable law and judicial decisions as subsidiary means of determining legal rules.
The graph covers selected materials from these categories together with UN resolutions linked to the collected court documents through citations.

The collection contains \num{6{,}396} ICJ documents and \num{505} documents from other courts and legal bodies with partial coverage of those institutions.
Its \num{370} legal instruments contain \num{12{,}375} provisions and include the four instruments governing the Court.\footnote{The UN Charter, the ICJ Statute, the Rules of Court and the Practice Directions.}
The \num{20{,}291} UN resolutions comprise \num{17{,}463} General Assembly resolutions and \num{2{,}828} Security Council resolutions.

\subsection{ICJ Graph Schema and Data Specification}
\label{subsec:icj_schema}
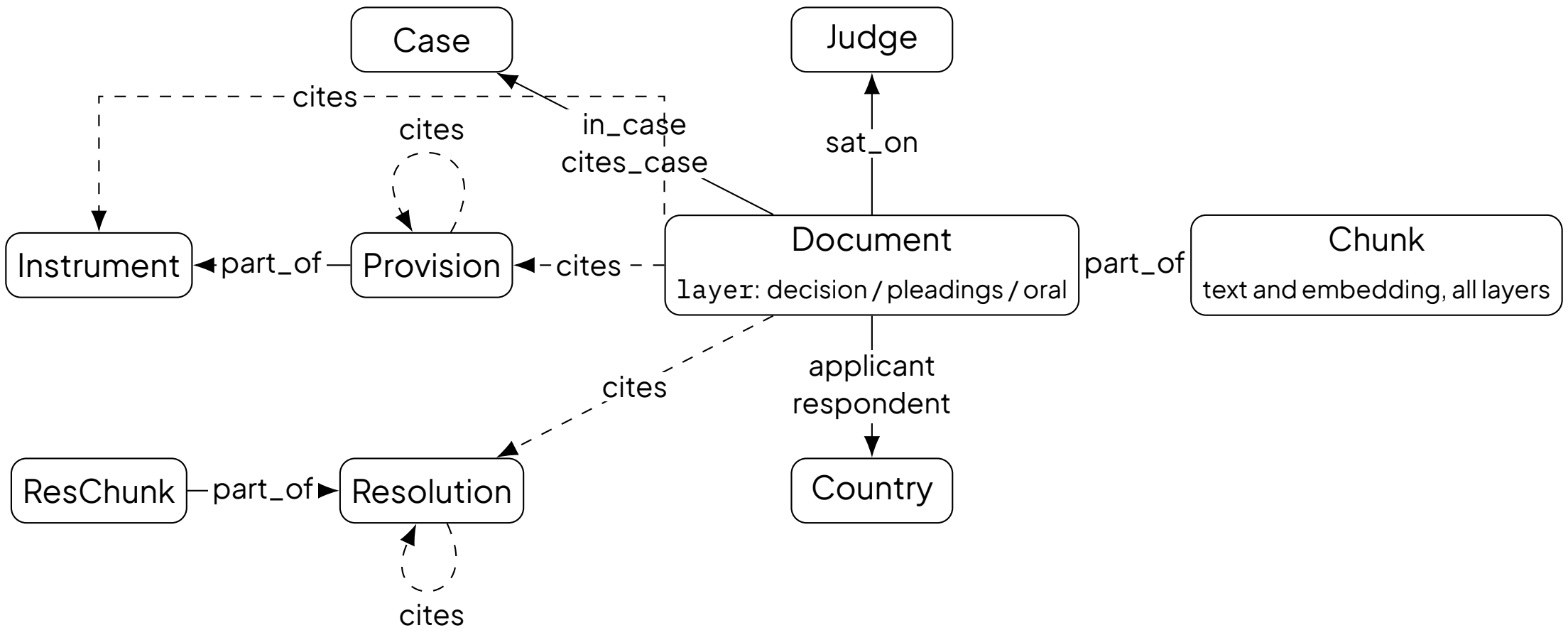
\begin{figure*}[!ht]
\centering
\begin{tikzpicture}[
  every node/.style={font=\small},
  ent/.style={draw, rounded corners, minimum height=6mm, minimum width=15mm,
              inner sep=3pt, align=center},
  rel/.style={-{Latex[length=2mm]}, font=\footnotesize},
  lbl/.style={font=\footnotesize, midway, fill=white, inner sep=1pt}
]
  \node[ent, minimum width=38mm] (doc) at (0,0)
        {Document\\[1pt]{\scriptsize \texttt{layer}: decision / pleadings / oral}};
  \node[ent, minimum width=30mm] (chunk) at (4.7,0)
        {Chunk\\[1pt]{\scriptsize text and embedding, all layers}};
  \node[ent] (judge) at (0,2.1)        {Judge};
  \node[ent] (ctry) at (0,-2.1)        {Country};
  \node[ent] (case) at (-4.1,2.1)      {Case};
  \node[ent] (prov) at (-4.1,0)        {Provision};
  \node[ent] (res) at (-4.1,-2.1)      {Resolution};
  \node[ent] (inst) at (-7.2,0)        {Instrument};
  \node[ent] (reschunk) at (-7.2,-2.1) {ResChunk};

  \draw[rel] (chunk) -- node[lbl]{\textsc{part\_of}} (doc);
  \draw[rel] (doc) -- node[lbl]{\textsc{sat\_on}} (judge);
  \draw[rel] (doc) -- node[lbl,align=center]{\textsc{applicant}\\\textsc{respondent}} (ctry);
  \draw[rel] (doc) -- node[lbl,align=center]{\textsc{in\_case}\\\textsc{cites\_case}} (case);
  \draw[rel,dashed] (doc) -- node[lbl]{\textsc{cites}} (prov);
  \draw[rel,dashed] (doc.north west) -- ++(0,1.1) -| node[lbl,pos=0.3]{\textsc{cites}} (inst.north);
  \draw[rel,dashed] (doc) -- node[lbl]{\textsc{cites}} (res);
  \draw[rel] (prov) -- node[lbl]{\textsc{part\_of}} (inst);
  \draw[rel] (reschunk) -- node[lbl]{\textsc{part\_of}} (res);
  \draw[rel,dashed] (prov) to[out=60,in=120,loop] node[lbl,above=1mm]{\textsc{cites}} (prov);
  \draw[rel,dashed] (res) to[out=-65,in=-115,loop] node[lbl,below=1mm]{\textsc{cites}} (res);
\end{tikzpicture}
\caption{Graph schema. Documents connect proceedings to cited authorities and carry party and bench relations where available.
ICJ documents are organized into decision, pleadings and oral layers.
Chunks hold document passages, provisions belong to legal instruments and ResChunks hold UN resolution passages.
Dashed edges mark citations to instruments, provisions and resolutions.
The arrow from Document to Case carries both case membership and case citations.}
\label{fig:schema}
\end{figure*}

\paragraph{Nodes.} The graph schema has nine node types, as illustrated in Figure~\ref{fig:schema}.

\begin{compactenum}
    \item \textbf{Case}: a proceeding identified by its court and case identifier. ICJ proceedings use their General List number, assigned when a case is filed.
    
    \item \textbf{Document}: one published PDF: a judgment or advisory opinion, an individual opinion appended to one, a written \emph{pleading} (a party's filing, such as a memorial or application), or an \emph{oral record} (a verbatim transcript of a hearing). ICJ documents carry a \texttt{layer} value of decision, pleadings or oral. The \num{505} documents from other adjudicative bodies have no layer label.
    
    \item \textbf{Chunk}: a passage of a document, cut on its own unit where possible; every chunk also
    carries an embedding.
    
    \item \textbf{Provision}: an article or other numbered unit of a legal instrument.
    
    \item \textbf{Instrument}: the parent of its provisions.
    \item \textbf{Resolution}: a UN General Assembly or Security Council resolution, keyed by its canonical symbol and carrying available metadata and full text.
    \item \textbf{ResChunk}: an ordered passage of a resolution body, with an embedding for semantic retrieval.
    \item \textbf{Judge}: a judge who took part in hearing a case.
    \item \textbf{Country}: a party to the case.
\end{compactenum}

\paragraph{Edges.}
\begin{compactenum}
    \item \textsc{in\_case}: Document to the Case it was filed in.
    \item \textsc{cites\_case}: Document to a Case it cites. 
    \item \textsc{cites}: Document to Instrument, Provision or Resolution; Provision to Provision and Resolution to Resolution for cross-references.
    \item \textsc{part\_of}: Chunk to Document, Provision to Instrument and ResChunk to Resolution.
    \item \textsc{sat\_on}: Document to Judge. 
    \item \textsc{applicant}/\textsc{respondent}: Document to Country.
\end{compactenum}

In the graph, party and judge relations are attached to \textbf{Document} nodes. Retrieving a document therefore also returns any party and bench information stated in that document. Appendix~\ref{sec:appendix-schema} describes the complete schema and its coverage.

Each \textbf{Chunk} and \textbf{ResChunk} is embedded into a \num{4096}-dimensional vector using \texttt{Qwen3-Embedding-8B}\footnote{We embedded the text with a self-hosted Qwen3-Embedding-8B endpoint on a single Nvidia V100 32GB GPU using vLLM.}~\cite{zhang2025qwen3embeddingadvancingtext}, so both Court documents and resolution passages can be retrieved semantically.
We release the embedding code with its exact configuration.
Parquet tables store the vectors alongside text and metadata, with float16 precision represented as float32 lists.
A vector-free Neo4j dump preserves the graph structure and non-vector properties.
The accompanying runtime imports stored vectors by application identifier or generates vectors through a configurable embedding endpoint.

\subsection{Graph Construction}
\label{sec:construction}

We re-parse the \num{2{,}285} historical documents from CD-ICJ \citep{fobbe2023cdicj} into paragraphs and generate embeddings for each paragraph. 
The additional \num{122} decisions issued after that corpus's cutoff were gathered and processed directly from the Court's site. 
We collect written pleadings and oral records from the Court's case pages and retain their document-layer labels.

\paragraph{Document Discovery.}
Case pages at \url{icj-cij.org/case/{n}} are plain HTML and enumerate
every document as a structured block under a document-type section header. We classify
each PDF's \texttt{layer} from its header (Algorithm~\ref{alg:discovery}). 
Two edge cases keep this from breaking, both handled as Algorithm~\ref{alg:discovery} shows: matching a document to its case by integer folder number (leading zeros differ across eras) and a filename-then-title fallback for cases listed under one flat ``Latest Developments'' heading.

\begin{figure*}[!ht]
\centering
\begin{minipage}{0.92\textwidth}
\begin{algorithm}[H]
\caption{Layer discovery for one case page. Section keywords: written\slash institution give
\emph{pleadings}, oral gives \emph{oral}, judgment\slash order\slash advisory\slash opinion give
\emph{decision}, summary\slash press\slash correspondence give \emph{drop}. Filename type codes:
wri\slash app\slash req\slash int\slash spe give \emph{pleadings}, jud\slash ord\slash adv give \emph{decision}, ora\slash cor
give \emph{oral}, pre\slash sum\slash oth give \emph{drop}.}
\label{alg:discovery}
\small
\begin{algorithmic}[1]
\Require case number $n$
\State $R \gets \emptyset$
\State $H \gets$ HTML of \texttt{icj-cij.org/case/}$n$ \quad (plain HTTP, back off on 429)
\ForAll{document blocks $b$ in $H$}
  \State $s \gets$ nearest document-type header above $b$
  \ForAll{PDF links $(f, \mathit{label})$ in $b$}
    \State parse folder, type code and title from $f$ and $b$
    \If{folder $\neq n$ as integers}
      \State skip this link \quad (zero-padding differs by era; match by integer)
    \Else
      \State $\ell \gets$ layer from $s$; if unknown, from the type code, then the title
      \If{$\ell$ is decision, pleadings, or oral}
        \State add $(f, \ell)$ to $R$
      \EndIf
    \EndIf
  \EndFor
\EndFor
\State \Return $R$
\end{algorithmic}
\end{algorithm}
\end{minipage}
\end{figure*}

\paragraph{Text and structure.}
We extract readable PDF text with PyMuPDF and use optical character recognition (OCR) for scanned pages or unusable text layers.
Appendix~\ref{sec:appendix-pipeline} describes the text-quality checks and OCR coverage.

\emph{Decision} documents run a two-pass pipeline (OCR with semantic tags, then a line classifier) to recover their structure: \emph{composition} (the judge and parties), the \emph{headnote} (the pre-reasons summary), the body's nested sections and the \emph{dispositif} (the operative paragraphs, with each judge's vote).
Each chunk records its zone, separating reasoning from decision.

\emph{Pleadings} and \emph{Oral} documents use the same pipeline, without composition and dispositif extraction.
We divide pleadings by numbered paragraph and oral records by speaker turn using regular expressions that follow each document's numbering and speaker conventions.
Legacy scans with no detectable structure (\num{24}\% of pleadings, \num{17}\% of oral chunks) fall back to a fixed window chunking with an empty label.

\paragraph{Related court decisions and legal instruments.}
We identify decisions from other courts and tribunals through citations in the ICJ documents, then retrieve matching records from the relevant court websites and legal archives.
Sources include the UN Reports of International Arbitral Awards and the European Court of Human Rights' HUDOC database.
Matching uses the available case names, dates and citation identifiers, so coverage follows the cited decisions for which records can be retrieved.
For legal instruments, we combine the Court's governing instruments with treaties identified in the collected documents.
The acquisition pipeline searches the UN Treaty Collection by title and uses treaty-series records or other legal repositories to retrieve the text.
Treaty texts reproduced in pleadings provide an additional source where a separate text is unavailable.
The pipeline divides the available texts into numbered provisions and retains their links to the parent instrument.

\begin{figure}[!t]
\centering
\resizebox{\columnwidth}{!}{%
\begin{tikzpicture}[
  node distance=6mm,
  every node/.style={font=\footnotesize},
  box/.style={draw, rounded corners, align=center, inner sep=3pt,
              text width=52mm, font=\footnotesize},
  test/.style={box, fill=black!5},
  res/.style={draw, rounded corners, align=center, inner sep=3pt,
              text width=15mm, font=\scriptsize},
  flow/.style={-{Latex[length=1.8mm]}},
  side/.style={flow, font=\scriptsize}
]
  \node[box] (hint) {\textbf{hint}: subject + parties + reporter year\\[1pt]
        {\scriptsize ``Nuclear Tests'', ``Australia v.\ France'', I.C.J. Reports 1974}};
  \node[test, below=of hint] (gate) {\textbf{year gate}\\[1pt]
        {\scriptsize keep cases whose active span, $\pm1$ year, contains 1974}};
  \node[test, below=of gate] (form) {\textbf{next name form}\\[1pt]
        {\scriptsize 1. subject + parties \quad 2. subject alone}};
  \node[test, below=of form] (exact) {\textbf{exact normalized name?}};
  \node[test, below=of exact] (score) {\textbf{fuzzy name score} (top, runner-up)};
  \node[test, below=of score] (thr) {\textbf{top} $\geq 0.90$?};
  \node[test, below=of thr] (tie) {\textbf{runner-up} $\geq 0.90$ \textbf{and} gap $< 0.07$?};
  \node[res, below=5.5mm of tie, fill=black!5] (link) {link\\{\scriptsize conf.\ = top}};

  \node[res, right=20mm of exact] (exlink) {link\\{\scriptsize conf.\ 1.0}};
  \node[res, right=20mm of form]  (none)   {no edge};

  \draw[flow] (hint) -- (gate);
  \draw[flow] (gate) -- (form);
  \draw[flow] (form) -- (exact);
  \draw[flow] (exact) -- node[right, font=\scriptsize, pos=0.45] {no} (score);
  \draw[flow] (score) -- (thr);
  \draw[flow] (thr) -- node[right, font=\scriptsize, pos=0.45] {yes} (tie);
  \draw[flow] (tie) -- node[right, font=\scriptsize, pos=0.4] {no} (link);
  \draw[side] (exact) -- node[above, font=\scriptsize] {yes} (exlink);
  \draw[side] (form) -- node[above, font=\scriptsize] {none left} (none);
  \draw[side] (thr.west) -- node[above, font=\scriptsize, inner sep=1.5pt] {no, too weak} ++(-32mm,0)
        |- ([yshift=-1.4mm]form.west);
  \draw[side] (tie.west) -- node[above, font=\scriptsize, inner sep=1.5pt] {yes, ambiguous} ++(-40mm,0)
        |- ([yshift=1.4mm]form.west);
\end{tikzpicture}}
\caption{Resolving one case-to-case citation. A hint names its case in two forms, tried in
order; each runs the same test and a form that fails either guard yields to the next.
Thresholds: link at \num{0.90}; abstain when the runner-up also clears \num{0.90} within
\num{0.07}.}
\label{fig:resolve}
\end{figure}
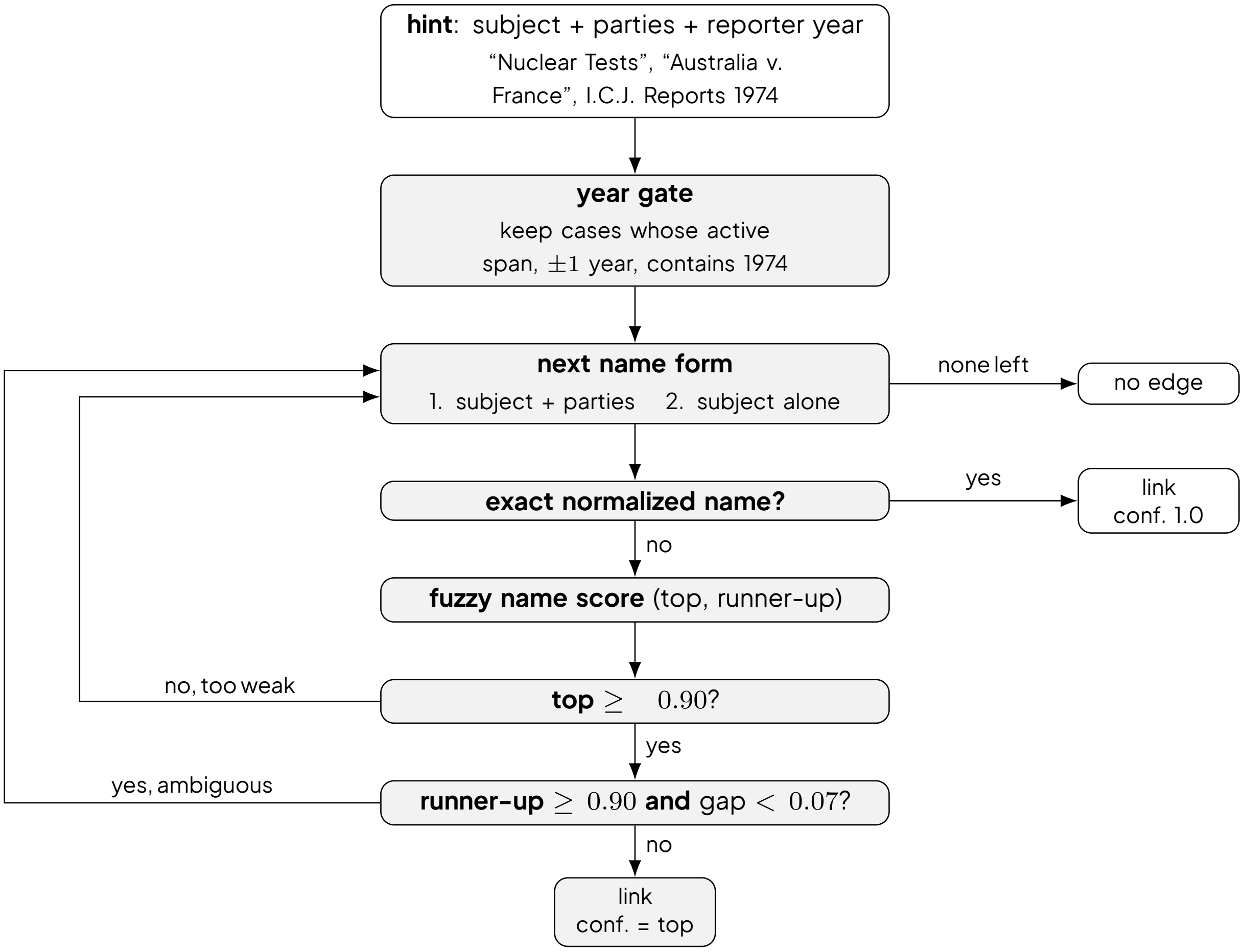

\paragraph{Case citation resolution.}
To extract decision-to-case citation relations for our ICJ graph, we first extract each citation mention as an unmapped record, which we refer to as a \emph{hint}. Each hint contains the case subject, parties and cited date or reporter year and is resolved using the pipeline shown in Figure~\ref{fig:resolve}.

We first apply a \textbf{year gate}, retaining candidate cases whose active span contains the cited year with a $\pm1$-year tolerance. We then try two name forms from the hint in order: \emph{subject + parties} followed by \emph{subject alone}.\footnote{An ICJ case usually has an official name of the form \emph{Subject (X v.\ Y)}, allowing the subject and parties to be extracted using regular expressions.} For each form, we first attempt an exact normalized-name match against the remaining candidates. If no exact match exists, we use fuzzy matching. A candidate is linked when the best fuzzy-match score is $\geq 0.90$, unless the runner-up also scores $\geq 0.90$ and lies within $0.07$ of the best match, in which case the result is treated as ambiguous and the next name form is tried. If neither form produces an unambiguous match, no citation edge is created, favoring missing edges over incorrect links.

\paragraph{Provision citation resolution.}
Citations to a provision are resolved by deterministic pattern matching rather than the
case pipeline above. A set of regular expressions finds each reference to an article,
chapter or rule in the text (for example \emph{Article 36} or \emph{Chapter VII}) and maps
it to the provision of the instrument named alongside it.
\paragraph{UN resolution sources and acquisition.}
We combine existing resolution texts with documents downloaded from the UN's official document service.
Security Council texts and metadata are imported from the Corpus of Resolutions: UN Security Council (CR-UNSC)~\citep{fobbe_2025_15154519}, using its English text and metadata files.
For General Assembly resolutions, we import English texts from the UNxml collection~\citep{unxml_ga}.
The PDF acquisition pipeline requests English documents by resolution symbol from \url{https://documents.un.org}, with alternative UN document links for older records.
We extract text directly from PDFs where possible and apply OCR to scanned documents.
When a PDF contains several General Assembly resolutions, we isolate the text for the requested resolution before dividing it into passages.
Records from these sources are joined by normalized resolution symbol.

\paragraph{UN-resolution citation resolution.}
Candidate mentions are detected from forms such as \emph{Security Council resolution 1244 (1999)}, \emph{resolution 2625 (XXV)} and canonical symbols.
The pipeline normalizes accepted mentions to General Assembly or Security Council symbols and links the citing document to the corresponding \textbf{Resolution} node.
Resolution texts are segmented into ordered \textbf{ResChunk} nodes; references found inside a resolution create Resolution-to-Resolution edges.
Resolution metadata supports filtering and canonical symbols identify individual resolutions.
Title and passage embeddings support searches by meaning.

\paragraph{Edge validation.}
We manually checked a random sample of \num{250} edges, half document-to-case and half document-to-provision, reading each citing document against the sentence the extractor matched. 
We verified \num{234} of the \num{250} edges against the citing text. 
The remaining \num{16} edges could not be verified because the citing text used ambiguous references such as ``the \num{1996} Judgment'' without a case name. 
Counting those \num{16} as errors, precision is \num{0.94} (\num{95}\% confidence interval \num{0.90} to \num{0.96}): all \num{125} provision edges held and \num{109} of the \num{125} case edges. 

This audit measures precision for sampled case and provision edges; it does not measure citation recall or validate UN-resolution edges.
Table~\ref{tab:stats} reports the ICJ document layers, while Appendix~\ref{sec:appendix-schema} reports node and relationship counts for the complete graph.

\begin{table}[!ht]
\centering
\small
\begin{tabular}{lrrr}
\toprule
 & Decision & Pleadings & Oral \\
\midrule
Cases covered      & \num{199}       & \num{194}       & \num{163} \\
Documents          & \num{2{,}407}   & \num{2{,}114}   & \num{1{,}875} \\
Chunks (passages)  & \num{121{,}438} & \num{682{,}547} & \num{291{,}669} \\
Pages              & \num{34{,}526}  & \num{336{,}059} & \num{118{,}346} \\
\bottomrule
\end{tabular}
\caption{ICJ document coverage by layer. Document, case and chunk counts are computed from the released Parquet tables; page counts describe the source PDFs.
The ICJ site lists \num{1{,}852} pleadings and \num{1{,}189} oral records across the \num{199} cases with decisions.
Our document counts refer to individual PDFs, so a listed filing with several volumes can contribute multiple documents.
The ICJ layers contain English and bilingual editions.}
\label{tab:stats}
\end{table}
\section{Citation-Prediction Benchmark}
\label{sec:baseline-setup}
We evaluate whether ICJ-Graph helps language-model agents identify authorities that the Court later cites.
For each available case numbered \num{001}--\num{200}, we select one target decision, yielding \num{199} targets\footnote{Case \num{002} was excluded as it is absent from the corpus}.
The selector prefers the latest judgment or advisory opinion; when neither exists, it uses the latest primary order.
This yields a broad historical benchmark spanning the available corpus.

\subsection{Task Construction}
\label{subsec:exp_setups}
\paragraph{Model-visible input.}
Each example supplies the case identifier, name, type, procedural stage and year.
For contentious cases it also supplies the parties and their claims; for advisory proceedings it supplies the requesting body and legal questions.
The model predicts three exact identifier sets: cited ICJ case identifiers, cited provision keys and cited UN-resolution symbols.
The agent also returns case-type-specific outcome fields for later analysis, but the present evaluation reports only the three citation tasks.

\paragraph{Temporal graph contract.}
Each target is evaluated against a graph view fixed immediately before its decision date.
Only documents with an \texttt{issued\_date} strictly earlier than the target date are visible.
The target document, documents issued on the same day, later documents and documents without a usable date are excluded.
The same contract is applied to graph traversal and semantic retrieval.
This prevents direct access to the held-out decision and its extracted citation edges, although it cannot remove target knowledge already present in a model's pretraining data.

\paragraph{Access conditions.}
We compare three complete agent configurations:
\begin{compactenum}
    \item \textbf{Parametric}: no external tools.
    \item \textbf{Cypher + Semantic}: \texttt{graph\_schema}, read-only \texttt{cypher\_read} and \texttt{search\_content}.
    \item \textbf{Designed Tools Only}: task-specific interfaces for cases, provisions, documents, citations, judges and resolutions. The complete list appears in Appendix~\ref{sec:appendix-tools}.
\end{compactenum}
The tool conditions apply a grounding rule: a predicted provision or resolution must have appeared in tool output and case identifiers must be supported by retrieved graph evidence.
The Parametric condition has no corresponding evidence channel.
Consequently, Parametric-versus-tool comparisons measure complete research configurations rather than a pure causal effect of tool access.
The Designed-versus-Cypher comparison is the more controlled test of interface design because both conditions use the same evidence policy and temporal graph.

\paragraph{Models and repetitions.}
We evaluate \texttt{gpt-5.6-luna}, \texttt{deepseek-v4-flash-0731}, \texttt{deepseek-v4.1-flash}, \texttt{muse-spark-1.3} and \texttt{gpt-5-mini}.
Every model--condition pair is run three times on all \num{199} targets, for \num{8{,}955} model--target runs in total.
Each run uses the same structured output schema and the same research-step limit.

\paragraph{Metrics and baselines.}
For each citation family, we compute exact-set precision, recall and F1 by pooling true positives, false positives and false negatives across the \num{199} targets within a run.
Tables report the mean and sample standard deviation across the three runs.
Query-independent baselines return the globally most frequently cited authorities: \(k\in\{6,18\}\) for cases, \(k\in\{5,12\}\) for provisions and \(k\in\{1,2\}\) for UN resolutions.
These baselines measure how much performance is available from citation popularity alone.

\clearpage
\section{Results}
\label{subsec:results}
\noindent\begin{minipage}{\textwidth}
\captionsetup{type=table}
\centering
\small
\setlength{\tabcolsep}{5pt}
\begin{tabular}{llccc}
\toprule
\textbf{Model} & \textbf{Condition} & \textbf{Case F1} & \textbf{Provision F1} & \textbf{UN-resolution F1} \\
\midrule
\multirow{3}{*}{\texttt{gpt-5.6-luna}}
 & Parametric & $.334 \pm .004$ & $.228 \pm .004$ & $.198 \pm .017$ \\
 & Cypher + Semantic & $.356 \pm .007$ & $\mathbf{.302 \pm .006}$ & $\mathbf{.266 \pm .034}$ \\
 & Designed Tools & $\mathbf{.378 \pm .008}$ & $.300 \pm .009$ & $\mathbf{.266 \pm .024}$ \\
\midrule
\multirow{3}{*}{\texttt{deepseek-v4-flash-0731}}
 & Parametric & $.277 \pm .015$ & $.232 \pm .005$ & $.164 \pm .019$ \\
 & Cypher + Semantic & $.348 \pm .014$ & $.273 \pm .037$ & $.289 \pm .037$ \\
 & Designed Tools & $\mathbf{.388 \pm .015}$ & $\mathbf{.324 \pm .003}$ & $\mathbf{.348 \pm .018}$ \\
\midrule
\multirow{3}{*}{\texttt{deepseek-v4.1-flash}}
 & Parametric & $.341 \pm .006$ & $.272 \pm .004$ & $.259 \pm .013$ \\
 & Cypher + Semantic & $.359 \pm .021$ & $.301 \pm .005$ & $.294 \pm .037$ \\
 & Designed Tools & $\mathbf{.395 \pm .009}$ & $\mathbf{.371 \pm .016}$ & $\mathbf{.357 \pm .037}$ \\
\midrule
\multirow{3}{*}{\texttt{muse-spark-1.3}}
 & Parametric & $.372 \pm .009$ & $.325 \pm .003$ & $.263 \pm .007$ \\
 & Cypher + Semantic & $.397 \pm .006$ & $.331 \pm .006$ & $\mathbf{.359 \pm .022}$ \\
 & Designed Tools & $\mathbf{.415 \pm .005}$ & $\mathbf{.349 \pm .006}$ & $.355 \pm .009$ \\
\midrule
\multirow{3}{*}{\texttt{gpt-5-mini}}
 & Parametric & $.146 \pm .010$ & $.147 \pm .007$ & $.085 \pm .010$ \\
 & Cypher + Semantic & $.172 \pm .010$ & $.166 \pm .013$ & $.076 \pm .005$ \\
 & Designed Tools & $\mathbf{.222 \pm .007}$ & $\mathbf{.200 \pm .007}$ & $\mathbf{.104 \pm .012}$ \\
\bottomrule
\end{tabular}
\caption{Micro-F1 over \num{199} target decisions, reported as mean $\pm$ sample standard deviation across three runs. Bold marks the best condition for each model and citation family; tied values are both bold.}
\label{tab:main-results}
\end{minipage}\par\medskip

\noindent\begin{minipage}{\textwidth}
\captionsetup{type=table}
\centering
\small
\begin{tabular}{lcc}
\toprule
\textbf{Citation family} & \textbf{Small \(k\)} & \textbf{Large \(k\)} \\
\midrule
Case & $.138$ (\(k=6\)) & $.179$ (\(k=18\)) \\
Provision & $.158$ (\(k=5\)) & $.232$ (\(k=12\)) \\
UN resolution & $.013$ (\(k=1\)) & $.024$ (\(k=2\)) \\
\bottomrule
\end{tabular}
\caption{Micro-F1 of query-independent most-frequent-citation baselines.}
\label{tab:frequency-baselines}
\end{minipage}\par\medskip

Designed Tools achieves higher mean F1 than answering without tools (Parametric) for all five models across case, provision and UN-resolution citations.
Compared with Cypher + Semantic, it gives every model a higher case-citation score and improves most provision and UN-resolution scores.
The exceptions are a provision-score decrease from \num{0.302} to \num{0.300} for \texttt{gpt-5.6-luna} and a UN-resolution decrease from \num{0.359} to \num{0.355} for \texttt{muse-spark-1.3}.
For \texttt{gpt-5.6-luna}, both methods have the same reported UN-resolution F1 of \num{0.266}.

The highest case-citation F1 is \num{0.415}, achieved by \texttt{muse-spark-1.3} with Designed Tools.
For provision citations, \texttt{deepseek-v4.1-flash} with Designed Tools scores highest at \num{0.371}.
For UN-resolution citations, \texttt{muse-spark-1.3} with Cypher + Semantic has the highest mean F1 at \num{0.359}, closely followed by \texttt{deepseek-v4.1-flash} with Designed Tools at \num{0.357}.

\section{Discussion}
\label{sec:discussion}
Case-citation prediction shows the most consistent gains from tools designed for citation research, with higher mean F1 than general graph queries and text search for every model.
For provisions and UN resolutions, the relative performance of the two tool configurations varies by model.
These comparisons show how the choice of tools affects performance within the tested configurations, while differences in search and citation-selection behavior remain possible explanations for the observed gains.

\section{Conclusion}
\label{sec:conclusion}
We introduce ICJ-Graph, a structured corpus connecting ICJ cases and documents to cited precedents, legal provisions and UN resolutions, with support for graph traversal and semantic retrieval.
A \num{199}-target citation-prediction benchmark illustrates one use of the resource across case, provision and resolution citations.

\section*{Availability}
ICJ-Graph is hosted at~\url{https://huggingface.co/datasets/VISAI-AI/icj-citation-graph} with a Neo4j dump without embeddings and Parquet tables containing stored embeddings.
The accompanying Docker Compose setup supports graph restoration and vector import or generation.
The benchmark is publicly available at~\url{https://huggingface.co/datasets/VISAI-AI/icj-precedent-cite-bench} with \num{199} test cases, reference citations, \num{8955} model predictions and scoring code.

\section*{Limitations}
\label{sec:limitations}
\paragraph{Corpus and labels.}
Coverage of the ICJ, other adjudicative bodies and UN resolutions is partial and follows the available source records.
Missing, revised or imperfectly digitized documents can produce gaps in the graph.
Automated extraction can omit or misresolve parties, judges, provisions, cases and resolution symbols.
The benchmark gold sets are derived from these extracted citations rather than exhaustive per-decision legal annotation.
The scores measure agreement with the extracted references and give no credit to other potentially relevant sources. They do not assess the quality of the model's legal reasoning.

\paragraph{Temporal gating does not remove pretraining exposure.}
The graph contract prevents a model from retrieving the target, same-day, later, or undated documents during a run.
Many historical decisions may nevertheless have appeared in model pretraining data, so Parametric performance cannot be interpreted as prediction from the supplied facts alone.

\paragraph{Conditions combine access and evidence policy.}
Tool-enabled agents must ground predicted identifiers in retrieved evidence, whereas the Parametric agent has no retrieval channel.
The comparison therefore measures end-to-end configurations and does not isolate the causal effect of tool access.
Designed Tools versus Cypher + Semantic is more controlled, but tool names, result shapes and the model's planning behavior still change together.

\paragraph{Models and repetitions.}
The benchmark covers five models and three repeated runs per configuration.
The repeated runs characterize stochastic variation within this setup; they are not independent samples of cases.
Results may change with other models, prompts, budgets, tool implementations, or graph versions.

\paragraph{Outcome fields are not evaluated here.}
The run schema includes jurisdiction, competence, discretion and free-text conclusions appropriate to each case type.
This paper reports only citation prediction because a defensible common scoring rule for those heterogeneous legal outcomes has not yet been established.

\paragraph{Reproducibility.}
The distributed corpus and runtime do not include the \num{199}-target evaluation artifacts needed to reproduce the reported scores.

\bibliography{references}
\appendix
\section{Graph Schema in Full}
\label{sec:appendix-schema}

The graph contains \num{1{,}416{,}857} nodes and \num{1{,}493{,}118} relationships.
Tables~\ref{tab:nodes} and~\ref{tab:edges} report counts from the exported Neo4j dump, checked against the source graph during export validation.
The \textsc{cites} relationship is separated by endpoint type because it represents citations to instruments, individual provisions and resolutions, as well as cross-references within those authorities.

\begin{table}[!ht]
\centering
\small
\begin{tabular}{lr}
\toprule
Node & Count \\
\midrule
Chunk       & \num{1{,}206{,}400} \\
ResChunk    & \num{169{,}617} \\
Resolution  & \num{20{,}291} \\
Provision   & \num{12{,}375} \\
Document    & \num{6{,}901} \\
Case        & \num{571} \\
Instrument  & \num{370} \\
Judge       & \num{220} \\
Country     & \num{112} \\
\bottomrule
\end{tabular}
\caption{Nodes in the complete graph. Case nodes comprise \num{200} ICJ proceedings and \num{371} proceedings from other adjudicative bodies.
The ICJ decision layer covers \num{199} cases; case \num{201} is represented by pleadings.}
\label{tab:nodes}
\end{table}

\begin{table}[!ht]
\centering
\small
\begin{tabular}{llr}
\toprule
Edge & From $\to$ To & Count \\
\midrule
\textsc{cites\_case} & Document $\to$ Case       & \num{7{,}449} \\
\textsc{cites}       & Document $\to$ Instrument & \num{16{,}287} \\
\textsc{cites}       & Document $\to$ Provision  & \num{12{,}232} \\
\textsc{cites}       & Document $\to$ Resolution & \num{2{,}054} \\
\textsc{cites}       & Provision $\to$ Provision & \num{242} \\
\textsc{cites}       & Resolution $\to$ Resolution & \num{53{,}110} \\
\textsc{sat\_on}     & Document $\to$ Judge      & \num{5{,}288} \\
\textsc{respondent}  & Document $\to$ Country    & \num{586} \\
\textsc{applicant}   & Document $\to$ Country    & \num{577} \\
\textsc{in\_case}    & Document $\to$ Case       & \num{6{,}901} \\
\textsc{part\_of}    & Chunk $\to$ Document      & \num{1{,}206{,}400} \\
\textsc{part\_of}    & Provision $\to$ Instrument & \num{12{,}375} \\
\textsc{part\_of}    & ResChunk $\to$ Resolution & \num{169{,}617} \\
\bottomrule
\end{tabular}
\caption{Relationships in the complete graph, separated by endpoint type. Document citation counts include records from the ICJ and other adjudicative bodies.}
\label{tab:edges}
\end{table}

The Neo4j dump preserves all relationship types and excludes embeddings and vector indexes.
Parquet tables contain text, metadata, selected citation relationships and stored vectors, joined to graph nodes through stable application identifiers.
Neo4j internal identifiers are regenerated during export and are unsuitable as join keys.
Party and judge relations are retained in the dump where available and do not cover every document.

All document chunks, resolution chunks and provisions have stored \num{4096}-dimensional Qwen3-Embedding-8B vectors.
Case-name vectors are available for \num{427} of \num{571} cases and resolution-title vectors for \num{17{,}649} of \num{20{,}291} resolutions.
The runtime supports vector import from Parquet and generation through a configurable embedding endpoint, with vector indexes created after vectors are loaded.

\section{Language by Layer}
\label{sec:appendix-lang}

The corpus is not uniformly one language and the split follows how the Court publishes
each kind of document rather than any choice of ours. Table~\ref{tab:lang} gives the
language labels of the ICJ documents. Decisions are
issued as separate English and French versions, of which we take the English, so the
decision layer is monolingual. Verbatim records of hearings are published as a single
volume, so the oral layer is almost entirely \emph{bilingual} in the Court's sense of the
word. Pleadings sit between the two, mostly English with a bilingual minority.

\emph{Bilingual} describes the edition rather than the languages in its text, a distinction
that matters when filtering documents. The Court publishes a hearing as a ``bilingual
version'' offered under an \emph{Original Language} link, meaning the volume preserves
whatever language each speaker used rather than translating it. A hearing argued entirely
in English therefore yields an all-English bilingual volume. Classifying the released
text of the \num{1{,}874} bilingual oral records by function-word frequency, sampled at
even intervals through each document, gives \num{499} English-dominant, \num{272}
French-dominant and \num{432} carrying substantial argument in both; the rest hold too
little text to call. Retrieval over the oral layer therefore needs to accommodate both English and French passages.

Table~\ref{tab:lang} therefore reports the label the Court assigns, not a measurement of
the text. The source labels do not always match the body text: \num{9} documents labelled English contain French text.

\begin{table}[t]
\centering
\small
\begin{tabular}{lrrr}
\toprule
Layer & English & Bilingual & Total \\
\midrule
Decision  & \num{2{,}407} & ---           & \num{2{,}407} \\
Pleadings & \num{2{,}046} & \num{68}      & \num{2{,}114} \\
Oral      & \num{1}       & \num{1{,}874} & \num{1{,}875} \\
\midrule
Total     & \num{4{,}454} & \num{1{,}942} & \num{6{,}396} \\
\bottomrule
\end{tabular}
\caption{Documents by language, counting a bilingual volume once.}
\label{tab:lang}
\end{table}

\section{ICJ Document Construction}
\label{sec:appendix-pipeline}

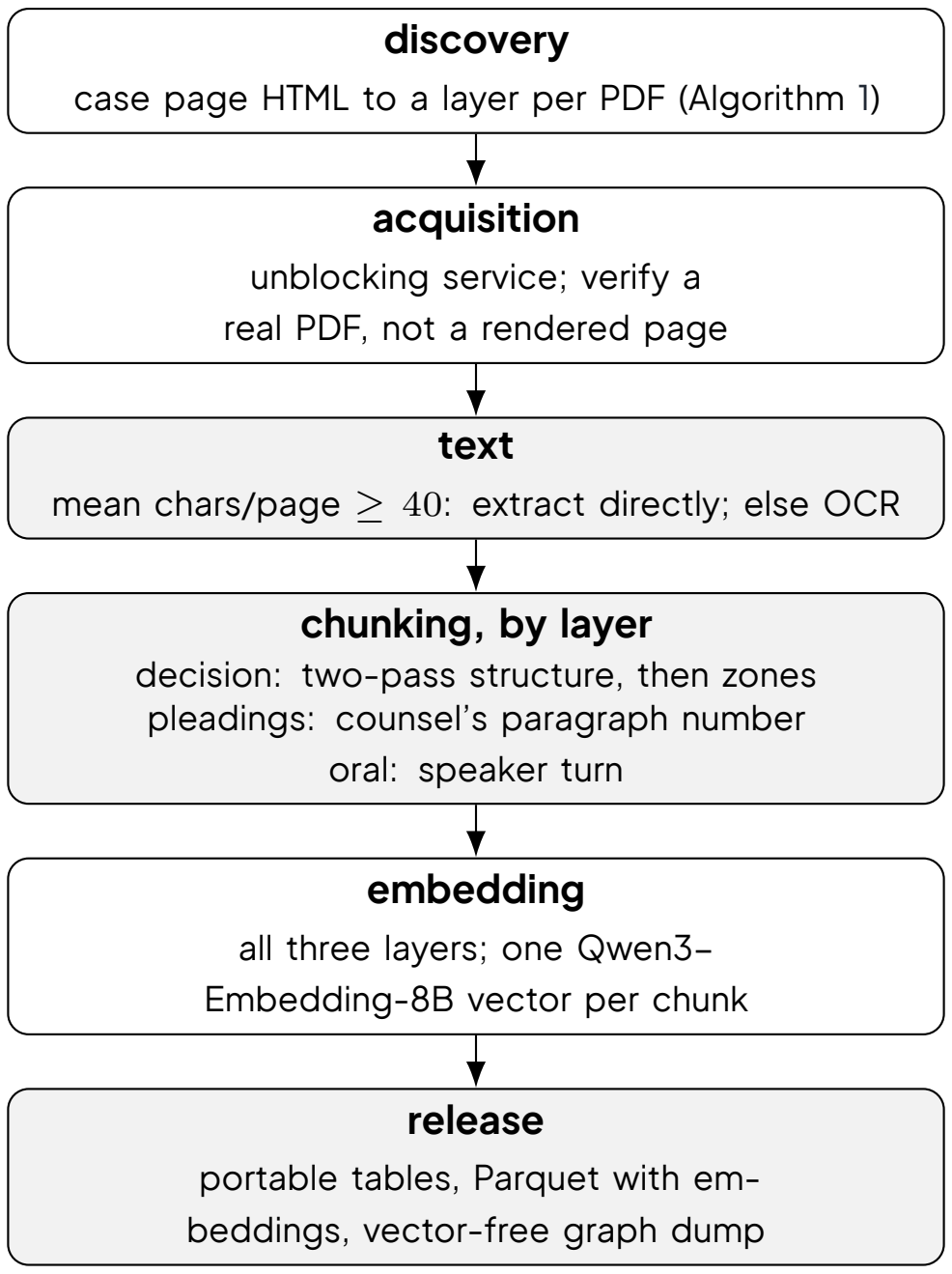
\begin{figure}[t]
\centering
\begin{tikzpicture}[
  node distance=3.6mm,
  stage/.style={draw, rounded corners, align=center, inner sep=3pt,
                text width=62mm, font=\footnotesize},
  split/.style={stage, fill=black!5},
  flow/.style={-{Latex[length=1.8mm]}}
]
  \node[stage] (disc) {\textbf{discovery}\\[1pt]{\scriptsize case page HTML to a layer
        per PDF (Algorithm~\ref{alg:discovery})}};
  \node[stage, below=of disc] (acq) {\textbf{acquisition}\\[1pt]{\scriptsize unblocking
        service; verify a real PDF, not a rendered page}};
  \node[split, below=of acq] (text) {\textbf{text}\\[1pt]{\scriptsize mean chars/page
        $\geq 40$: extract directly; else OCR}};
  \node[split, below=of text] (chunk) {\textbf{chunking, by layer}\\[1pt]{\scriptsize
        decision: two-pass structure, then zones\\
        pleadings: counsel's paragraph number\\
        oral: speaker turn}};
  \node[stage, below=of chunk] (emb) {\textbf{embedding}\\[1pt]{\scriptsize all three
        layers; one Qwen3-Embedding-8B vector per chunk}};
  \node[split, below=of emb] (out) {\textbf{release}\\[1pt]{\scriptsize portable tables,
        Parquet with embeddings, vector-free graph dump}};

  \draw[flow] (disc) -- (acq);
  \draw[flow] (acq) -- (text);
  \draw[flow] (text) -- (chunk);
  \draw[flow] (chunk) -- (emb);
  \draw[flow] (emb) -- (out);
\end{tikzpicture}
\caption{Construction, end to end. Chunking is the first layer-specific stage.}
\label{fig:pipeline}
\end{figure}

ICJ documents pass through discovery, acquisition, text extraction, chunking and embedding before export (Figure~\ref{fig:pipeline}).
The stages before chunking are layer-agnostic: discovery classifies a PDF's layer but
retrieves every layer the same way and the text test asks only whether a usable text
layer exists, which is a property of the file and not of what kind of document it
holds.

Chunking diverges by layer: decisions are cut on recovered structure, pleadings on
counsel's numbering, oral records on speaker turns. Embedding then covers all three layers,
one Qwen3-Embedding-8B vector per chunk.

We identify PDFs without a usable text layer by each document's mean characters per page,
taken over the whole document rather than its opening pages since recent filings front a
scanned cover onto a machine-readable body: below \num{40} the document is sent to OCR. A
page of prose runs to hundreds of characters and a scanned page to almost none, so the
cutoff sits in a wide gap, set by visual audit. This finds the \num{76} documents that are
scans. It cannot see a text layer that decodes to the wrong characters, which needs the
separate per-page text-quality detector, which sent \num{43}
documents garbled throughout for full re-OCR. So \num{119} of the \num{3{,}989} pleadings
and oral documents (\num{3.0}\%) have an OCR-derived text layer and a further \num{464}
keep their native text with the individual pages that detector flagged replaced by OCR. Every LLM stage of
construction (OCR, page classification and citation extraction) runs
\texttt{gemini-3.1-flash-lite} through a commercial API router.

\subsection{Benchmark Tool Conditions}
\label{sec:appendix-tools}
The benchmark exposes the following fixed capability sets:
\begin{compactenum}
\raggedright
    \item \textbf{Parametric}: The model answers using its existing knowledge without access to retrieval tools.
    \item \textbf{Cypher + Semantic}: The model can inspect the graph schema, write Cypher queries and search document passages by meaning. The available tools are \texttt{graph\_schema}, \texttt{cypher\_read} and \texttt{search\_content}.
    \item \textbf{Designed Tools Only}: The model uses predefined tools to find and read cases, legal provisions and UN resolutions without writing Cypher queries. The available tools are \texttt{list\_cases}, \texttt{get\_case}, \texttt{find\_provision}, \texttt{get\_provision}, \texttt{get\_document}, \texttt{citations}, \texttt{judges\_on\_case}, \texttt{cases\_by\_judge}, \texttt{list\_resolutions}, \texttt{find\_resolution}, \texttt{search\_resolution\_content} and \texttt{get\_resolution}.
\end{compactenum}
All tools are read-only and operate through the target-specific temporal graph contract.
\texttt{find\_resolution} performs semantic title search, whereas \texttt{get\_resolution} reads an exact canonical symbol.

\section{Prediction Schema}
\label{sec:appendix-prediction}
Predictions are validated against a strict structured-output schema.
Both case types return three citation arrays, as shown below.
Contentious cases additionally return a three-way jurisdiction field and claim-level conclusions; advisory cases return competence, optional discretion and question-level conclusions.
Unknown fields are rejected.
Case identifiers are normalized to three digits, provision keys to lowercase and resolution symbols to uppercase.
Only the citation arrays are scored in this paper.
{\footnotesize
\begin{verbatim}
{
  "case_type": "contentious",
  "jurisdiction": "upheld|declined|mixed|null",
  "conclusions": [...],
  "citation_case_ids": ["NNN", ...],
  "citation_provision_keys": ["instrument:article_N", ...],
  "citation_resolution_symbols": ["A/RES/...", "S/RES/...", ...]
}
\end{verbatim}
}

\section{Licensing and Ethics}
\label{sec:license}

Public access does not by itself settle the right to reuse these documents.
The Rules of Court make pleadings and oral records \emph{accessible to the public} at or after the opening of oral proceedings (\citealp{icj_rules}, Art.~53), but this rule concerns access rather than reuse.
The ICJ separately claims copyright over its website and publications (\citealp{icj_disclaimer}: ``All rights reserved'') and neither the ICJ nor the UN provides specific guidance on text and data mining.
Moreover, the rule that court-authored legal texts are not protected by copyright \citep{georgia2020pro} depends on the court being the author and therefore does not clearly extend to pleadings written by the parties.
We consequently rely on the research-use, fair-use and public-interest rationales applied to legal-text corpora without claiming that these rationales conclusively resolve the rights status of the source documents.

The corpus also contains unscreened personal data.
The records identify agents, counsel, judges and other individuals; include testimony given under cross-examination; and, in cases concerning alleged atrocities, may describe identified individuals in detail.
We have not screened or redacted this information and do not claim that its release is low-risk.
We note only that the material was already published by the Court as part of the public record and that we reproduce it without alteration.

We license only the components that we created.
Our parsing, chunking, embeddings, citation graph and tooling are released under \texttt{CC-BY-4.0}, while decision documents inherited from CD-ICJ retain their upstream \texttt{CC0-1.0} designation.
We grant no licence over the underlying ICJ documents, other adjudicative records, legal instruments or UN resolutions; their source-specific terms continue to apply.
For material obtained from the ICJ website, we attach a usage request modeled on \texttt{CC-BY-NC-4.0}, asking users to credit both the Court and this dataset and to limit use to non-commercial purposes.
This request is not a Creative Commons licence and does not imply that the documents' rights status has been resolved.
We will honour takedown requests and maintain this position while seeking clarification from the Registry.

\fcend
\end{document}